\documentclass[11pt]{article}

\usepackage[utf8]{inputenc}
\usepackage[T1]{fontenc}
\usepackage{amsmath,amssymb,amsfonts}
\usepackage{booktabs}
\usepackage{graphicx}
\usepackage{hyperref}
\usepackage[margin=1in]{geometry}
\usepackage{enumitem}
\usepackage{array}
\usepackage{xcolor}
\usepackage{natbib}
\usepackage{microtype}
\usepackage{multirow}
\usepackage{tabularx}
\usepackage{caption}
\usepackage{subcaption}

\newcommand{\AuthorDisplayName}{Qingwei Lin}
\newcommand{\AuthorPdfName}{Qingwei Lin}
\newcommand{\AuthorAffiliation}{Independent Researcher}
\newcommand{\AuthorEmail}{1833340239@qq.com}

\hypersetup{
  colorlinks=true,
  linkcolor=blue,
  citecolor=blue,
  urlcolor=blue,
  pdftitle={Revisiting Auxiliary Losses for Conditional Depth Routing: An Empirical Study},
  pdfauthor={\AuthorPdfName}
}

\newcommand{\ie}{\textit{i.e.}}
\newcommand{\eg}{\textit{e.g.}}

\newcommand{\sg}{\mathrm{sg}}
\newcommand{\relu}{\mathrm{ReLU}}
\newcommand{\softplus}{\mathrm{softplus}}

\graphicspath{{figures/}}

\title{
  Revisiting Auxiliary Losses for Conditional Depth Routing:\\
  \large An Empirical Study
}

\author{
  \AuthorDisplayName\\
  \textit{\AuthorAffiliation}\\
  \texttt{\AuthorEmail}
}

\date{}

\begin{document}
\maketitle

\begin{abstract}
Conditional depth execution routes a subset of tokens through a lightweight \texttt{cheap FFN} while the remainder execute the standard \texttt{full FFN} at each controlled layer.
The central difficulty is \emph{gate training}: the gate decision must propagate through many layers before it influences the language modeling (LM) loss, so the resulting gradients are weak and noisy.
Auxiliary losses are commonly stacked to stabilise training, yet the interactions among them---particularly between a predictive auxiliary and explicit score supervision---have not been systematically compared under controlled conditions.

We evaluate two gate designs under a 157.5M-parameter decoder-only model with controller-only training, 50\% full-path budget, and 3-seed runs on a \texttt{fineweb-edu} subset.
The \textbf{MLP gate (G1)} maps the current hidden state directly to a utility score; the \textbf{JEPA-guided gate (G3)} introduces an action-conditional predictor that forecasts, in a low-dimensional latent space, the outcome of executing \texttt{full} versus \texttt{cheap} for each token, with alignment enforced against a fixed target projection head.
Under the standard recipe that includes oracle-style utility regression and pairwise rank supervision (jointly \textbf{util/rank}), G3 consistently improves early-to-mid optimisation dynamics relative to G1---lower average LM loss, faster threshold hits, ${\approx}\,10.3{\times}$ lower training gradient norms---in 3/3 seeds, with the 20k-step endpoint LM within a preset 0.005 heuristic reference threshold.
Ablations confirm that JEPA alignment is necessary for predictor non-collapse (A1) and that its weight is insensitive across $\lambda_{\text{JEPA}} \in \{0.25, 1.0, 2.0\}$ (A4).

A key finding emerges from ablation A3: simultaneously removing util/rank \textbf{improves} best/average LM and threshold-hit speed in 3/3 seeds for \emph{both} gate architectures; moreover, the early-to-mid optimisation advantage of G3 over G1 disappears, leaving the two gates indistinguishable on the reported LM metrics (though score-calibration behavior still differs).
We trace this to a structural mismatch: the oracle label assumes ``current layer forks; all subsequent layers execute full,'' whereas under actual gated execution each subsequent controlled layer routes only a fraction $\rho$ of its tokens through full, making the teacher off-policy.
Under controller-only training this bias can accumulate, rendering util/rank a \textbf{net-negative contribution under the current recipe} ($\lambda_{\text{util}} = 1.0$, $\lambda_{\text{rank}} = 0.2$).
As a byproduct, removing util/rank cuts the training-time FLOPs proxy from ${\approx}\,1.53{\times}$ to ${\approx}\,1.07{\times}$ the full-only baseline (measured wall-clock on a single V100-32GB: $2.87\mathrm{h}\!\to\!1.75\mathrm{h}$, ${\approx}\,39\%$ reduction) while simultaneously improving LM quality.
All conclusions are strictly scoped to the studied regime and are not extrapolated to larger scales, joint backbone training, or alternative oracle definitions.
\end{abstract}

\section{Introduction}
\label{sec:intro}

A substantial portion of the inference cost of modern language models arises from the full-activation paradigm, under which every token executes the complete feed-forward network (FFN) at every layer.
Conditional depth execution relaxes this constraint: at a subset of \emph{controlled layers}, some tokens execute the standard \texttt{full FFN} while the remainder execute a cheaper alternative (\texttt{cheap FFN}), thereby improving the compute--quality trade-off under a fixed budget.

The central difficulty of this framework resides not in the design of the cheap path, but in \textbf{gate/controller training}.
Although a budget loss can constrain the average full-path ratio, it cannot directly indicate whether a given token at a given layer warrants full computation.
When the gate relies solely on indirect gradients propagated from the LM loss through a long chain of layers, training tends to be slow and unstable.

To address this weak-supervision problem, prior work in conditional computation and routing has introduced various auxiliary signals to stabilize gate training, including budget/alive constraints~\citep{raposo2024mod}, router numerical-stability losses~\citep{zoph2022stmoe}, load-balancing losses~\citep{fedus2022switch}, and, in some cases, explicit router supervision or teacher-style signals.
However, the \emph{pairwise interactions} among these auxiliary signals in the specific setting of conditional depth routing---in particular, whether a predictive auxiliary and explicit score supervision are compatible, and whether their joint application is necessarily beneficial---have not, to our knowledge, been systematically compared under controlled conditions.

\paragraph{Scope and positioning.}
This paper evaluates a JEPA-inspired gate design for conditional depth routing and, through systematic ablation, discovers that the util/rank losses are net-negative under the tested recipe.
The main experiments run in a small, controlled regime (${\approx}\,157.5$M total parameters, ${\approx}\,0.83$--$0.85$M trainable controller parameters, with a supplementary \texttt{G1-costmatch} control at ${\approx}\,0.90$M; controller-only training, 50\% budget; a 25\% budget supplement is also reported).
Throughout, we use the shorthand ``\textbf{utility/rank}'' for the joint Huber-regression and pairwise-rank losses on the oracle label $u^{\star}$ (defined in~\S\ref{sec:methods:util}).
The two families of auxiliary signals under study are:

\begin{itemize}[leftmargin=*,itemsep=2pt]
\item \textbf{Predictive auxiliary (JEPA-inspired).}
We examine a JEPA-inspired variant that trains a predictor to forecast, in a low-dimensional latent space, the outcome of executing \texttt{full} versus \texttt{cheap} for each token at each controlled layer.
A fixed target projection head enforces alignment.
This auxiliary shapes the \emph{future features} on which the gate relies.
To our knowledge, instantiating JEPA's action-conditional latent-space prediction as an auxiliary supervision for conditional depth gates has not been systematically reported in the conditional computation literature; we treat it as a \emph{candidate auxiliary to be empirically evaluated}, not as a default baseline.

\item \textbf{Oracle supervision (utility/rank).}
During training, a counterfactual trajectory (``current layer forks into full vs.\ cheap; all subsequent layers execute full'') is run online to construct a token-level utility label $u^{\star} = \widetilde{\mathrm{CE}}_{\text{cheap}} - \widetilde{\mathrm{CE}}_{\text{full}}$.
Huber regression and pairwise rank losses then explicitly pull the gate output toward $u^{\star}$.
This auxiliary shapes the gate's \emph{output score itself}.
\end{itemize}

Intuitively, the two families supply stronger training signals from the ``feature'' and ``label'' directions, respectively, and combining them would appear to be at least as effective as using either in isolation.
We take this additivity expectation as our starting point and subject it to empirical scrutiny.

We organise findings along two main lines---the G1 vs.\ G3 comparison and the A3 ablation discovery---with supporting mechanistic evidence:

\begin{itemize}[leftmargin=*,itemsep=2pt]
\item \textbf{Main comparison (G1 vs.\ G3).}
Under the standard recipe with util/rank, the JEPA-guided gate (G3) consistently improves early-to-mid optimisation dynamics relative to the MLP gate (G1)---lower avg LM, faster threshold hits, ${\approx}\,10.3{\times}$ lower gradient norms---in 3/3 seeds, with the 20k-step endpoint LM within the 0.005 heuristic reference threshold.
JEPA alignment is necessary: without it, the predictor collapses (A1); its weight is insensitive in the range 0.25--2.0 (A4).

\item \textbf{Key ablation finding (A3).}
Simultaneously removing util/rank \textbf{improves LM in 3/3 seeds for both gate architectures}; moreover, the early-to-mid optimisation advantage of G3 over G1 disappears---per-seed LM differences fall within the 0.005 heuristic reference threshold, making the two gates indistinguishable on the reported LM metrics.
Both the direction and the stability of this result contradict the starting intuition that util/rank is a ``necessary auxiliary.''
We trace this to a structural mismatch: the oracle's ``subsequent-all-full'' assumption is off-policy relative to actual gated execution (\S\ref{sec:methods:util}).

\item \textbf{Supporting evidence (A2, 25\% budget).}
Token-wise semantic pairing provides measurable but secondary benefit (A2); the G3 vs.\ G1 endpoint LM falls within the 0.005 margin at 25\% budget as well, though with per-seed sign reversal.
\end{itemize}

The contributions of this paper are:

\begin{enumerate}[leftmargin=*,itemsep=2pt]
\item \textbf{A systematic evaluation of a JEPA-inspired gate for conditional depth routing.}
We compare the MLP gate (G1) and the JEPA-guided gate (G3) under identical conditions with 3-seed runs.
G3 improves optimisation dynamics under the standard recipe; ablations A1 and A4 establish that JEPA alignment is a necessary but weight-insensitive condition for predictor non-collapse.
As a minor observation, $\ell_2$ distance rather than cosine similarity is the appropriate collapse diagnostic for action-conditional predictors in this setup.

\item \textbf{A recipe-level discovery that oracle-style util/rank is net-negative.}
Under the current recipe ($\lambda_{\text{util}} = 1.0$, $\lambda_{\text{rank}} = 0.2$, ``subsequent-all-full'' oracle), removing util/rank jointly improves best/avg LM and threshold-hit speed in 3/3 seeds; furthermore, the early-to-mid optimisation advantage of G3 over G1 disappears, indicating that JEPA's LM benefit is at least partially coupled with the presence of util/rank.
We provide a retrospective diagnosis---the off-policy teacher bias---as a structurally grounded working hypothesis, while explicitly listing alternative explanations (H3--H5) that our data cannot yet exclude.
The distinction between ``feature shaping'' (JEPA) and ``score anchoring'' (util/rank) as two separable auxiliary roles emerges from this ablation.

\item \textbf{Quantified efficiency observation and explicit scope.}
Removing util/rank cuts the training-time FLOPs proxy from ${\approx}\,1.53{\times}$ to ${\approx}\,1.07{\times}$ the full-only baseline (measured wall-clock: ${\approx}\,39\%$ reduction), while inference-time compute is unchanged.
All conclusions are strictly bounded to the ${\approx}\,157.5$M / ${\approx}\,0.83$--$0.85$M trainable parameter, controller-only, 50\% budget regime on a \texttt{fineweb-edu} subset.
We do not extrapolate to larger scales, joint backbone training, downstream benchmarks, or production serving scenarios.
\end{enumerate}

\section{Related Work}
\label{sec:related}

\paragraph{Conditional computation and dynamic depth.}
Adaptive Computation Time (ACT)~\citep{graves2016adaptive} pioneered the idea of allocating different amounts of computation per token.
Depth-Adaptive Transformer~\citep{elbayad2020depth} and Confident Adaptive Language Modeling (CALM)~\citep{schuster2022confident} extended this to Transformer language models via learnable early-exit mechanisms that save computation at fixed quality.
LayerSkip~\citep{elhoushi2024layerskip} combines training-time early exit with inference-time self-speculative decoding.
Mixture-of-Depths (MoD)~\citep{raposo2024mod} introduces token-level routing within the Transformer, selecting a subset of tokens for full forward computation at each layer; the router training procedure is formally similar to the controller training studied in this work.
Rather than proposing a new routing paradigm, we evaluate a JEPA-inspired gate design and focus on a less directly examined question: \textbf{when multiple auxiliary losses are simultaneously introduced for gate training within controlled layers, what are their interactions?}

\paragraph{Mixture-of-Experts and router training.}
Starting from the sparsely-gated mixture-of-experts layer~\citep{shazeer2017moe} and its scaled realisations such as {GShard}~\citep{lepikhin2021gshard} and {GLaM}~\citep{du2022glam}, the MoE literature has developed an extensive discussion of router auxiliary losses: the load-balancing loss~\citep{fedus2022switch}, the router z-loss~\citep{zoph2022stmoe}, and expert-choice routing~\citep{zhou2022expertchoice} are all designed to stabilise router training and prevent collapse.
Our budget/alive losses are formally close relatives of these auxiliary losses, whereas our utility/rank losses more closely resemble teacher--student-style explicit distillation.
Consequently, our observations can also be read as evidence that \textbf{in routing-based conditional computation more broadly, the distribution mismatch between an oracle-style teacher and on-policy execution may be an underappreciated problem.}

\paragraph{Joint Embedding Predictive Architecture (JEPA).}
\citet{lecun2022path} proposed JEPA as a unified framework for learning world models---predicting the outcome of an action in latent space, with alignment between a predictor and a fixed target.
I-JEPA~\citep{assran2023ijepa} and V-JEPA~\citep{bardes2024vjepa} realized this framework in visual self-supervised learning.
In this paper, we adapt JEPA's action-conditional latent-space prediction to the conditional depth gate: \texttt{full} and \texttt{cheap} are treated as two candidate actions, and the predictor produces low-dimensional outcome summaries for each path, upon which a decision head scores utility.
We note that our JEPA variant does not constitute self-supervised representation learning in the strict sense---it serves as a \emph{predictive auxiliary} for gate training, with both the predictor and the decision head optimized under the joint LM loss, while the target projection head remains fixed without an EMA teacher.

\paragraph{Relationship between G3 and related router designs.}
To clarify the formal distinction between our G3 and existing routing/early-exit methods, we provide the following comparison:

\begin{table}[h]
\centering
\small
\renewcommand{\arraystretch}{1.08}
\begin{tabular}{@{}>{\raggedright\arraybackslash}p{0.20\textwidth}>{\raggedright\arraybackslash}p{0.35\textwidth}>{\raggedright\arraybackslash}p{0.37\textwidth}@{}}
\toprule
\textbf{Method} & \textbf{Explicit router supervision} & \textbf{Router output form} \\
\midrule
MoD & LM loss only & Scalar logit, top-$k$ selection \\
CALM & Self-supervised confidence head & Scalar halt probability \\
MoE balancing/z-loss & Numerical stability (not a policy teacher) & Expert router logits \\
\textbf{G3 + JEPA (ours)} &
Action-conditional latent prediction $+$ fixed-target alignment &
Two vectors $q_a \in \mathbb{R}^{d_s}$; decision head scores on top; top-$k$ selection \\
\bottomrule
\end{tabular}
\vspace{2pt}
\caption*{\footnotesize \textit{References.} MoD: \citet{raposo2024mod}. CALM: \citet{schuster2022confident}. MoE balancing/z-loss: \citet{fedus2022switch,zoph2022stmoe}.}
\end{table}

Thus G3 is formally closer to an ``action-conditional local world model,'' while utility/rank supervision corresponds to an oracle-style policy teacher.
\textbf{Placing these two families on the same platform and disentangling their effects is the specific contribution of this paper, rather than proposing a new routing paradigm.}

We scope our novelty claim precisely as follows: to our knowledge, the \emph{specific combination} of ``action-conditional latent-space prediction + fixed target alignment + a decision head based on differences between $q_{\text{full}}$ and $q_{\text{cheap}}$'' has not been systematically reported as auxiliary supervision for conditional depth gates.
The qualifier ``combination'' is essential: the MoD router can be described as a scalar form of action-conditional predictor, and the CALM confidence head can be viewed as a self-supervised head using later-layer consistency as signal---both share structural similarities with G3 at their respective levels of abstraction.
\textbf{We do not claim ``the first use of action-conditional prediction in routing''} at this broader level of abstraction, as MoD, CALM, and related works have precedent.
Our claim is narrower: the above three-element combination is used as auxiliary supervision \emph{alongside} an oracle-style policy teacher (utility/rank) on the same platform, yielding \emph{non-trivial interaction evidence} (\S\ref{sec:results:a3}).
The scientific contribution thus rests on: (a)~the specific combination; (b)~same-platform comparison; and (c)~empirical evidence of interactions with the oracle-style teacher---not on introducing a new routing paradigm.

\section{Methods}
\label{sec:methods}

\paragraph{Terminology (controller, gate, router).}
Throughout this paper, ``\textbf{controller}'' denotes the full trainable module that computes the token-level utility score $u_i$ (\ie, the MLP head in G1 or the predictor + decision head in G3), while ``\textbf{gate}'' and ``\textbf{router}'' are used interchangeably for the top-$k$ execution decision on top of $u_i$, following the conventions of~\citet{raposo2024mod} and~\citet{fedus2022switch} respectively.
Sentences of the form ``the gate learns $\ldots$'' therefore always refer to the controller's output, not to an additional trainable module.

\subsection{Conditional Depth Language Model and 50\% Budget}
\label{sec:methods:model}

Consider a model with $L$ decoder layers, of which the last $C$ layers are \emph{controlled layers}.
In each controlled layer~$i$, the attention sub-layer remains unchanged; the FFN sub-layer is replaced with a conditional execution structure.
For notational clarity and to maintain isomorphism with the implementation, we write out the attention residual sub-layer and the conditional FFN residual sub-layer explicitly.
Let $\mathbf{h}_i^{\mathrm{in}}$ denote the input to layer~$i$.
The attention residual sub-layer output is
\begin{equation}
  \mathbf{a}_i = \mathbf{h}_i^{\mathrm{in}} + \mathrm{Attn}\bigl(\mathrm{LN}_1(\mathbf{h}_i^{\mathrm{in}})\bigr).
\end{equation}
Within the same FFN residual sub-layer, the forward pass branches between \texttt{full} and \texttt{cheap} paths according to a token-level execution mask~$m_i$:
\begin{align}
  \mathbf{h}_i^{\text{full}} &= \mathbf{a}_i + \mathrm{FFN}_{\text{full}}\bigl(\mathrm{LN}_2(\mathbf{a}_i)\bigr), \\
  \mathbf{h}_i^{\text{cheap}} &= \mathbf{a}_i + \mathrm{FFN}_{\text{cheap}}\bigl(\mathrm{LN}_{2,\text{cheap}}(\mathbf{a}_i)\bigr), \\
  \mathbf{h}_i &= m_i \cdot \mathbf{h}_i^{\text{full}} + (1 - m_i) \cdot \mathbf{h}_i^{\text{cheap}},
\end{align}
where $m_i \in \{0, 1\}$ is the token-level execution mask (strictly binary at inference; replaced by the straight-through surrogate $m_i^{\mathrm{st}}$ during training to maintain a differentiable channel from LM loss to the gate, in the spirit of the straight-through estimator of \citet{bengio2013estimating} and the continuous relaxation used in \citet{jang2017gumbelsoftmax}).
The \texttt{full FFN} is the standard feed-forward network; the \texttt{cheap FFN} is a low-rank approximation:
\begin{equation}
  \mathrm{FFN}_{\text{cheap}}(\mathbf{x}) = \mathbf{W}_{\text{down}}\,\mathrm{SiLU}(\mathbf{W}_{\text{up}}\,\mathbf{x}),
  \qquad \mathbf{W}_{\text{up}} \in \mathbb{R}^{r \times d},\;
  \mathbf{W}_{\text{down}} \in \mathbb{R}^{d \times r},\;
  r \ll d_{\text{ff}}.
\end{equation}
Here $\mathrm{LN}_{2,\text{cheap}}$ is a dedicated LayerNorm for the cheap path, with parameters not shared with the full path's $\mathrm{LN}_2$.
$\mathbf{W}_{\text{down}}$ is initialized to all zeros, so that $\mathrm{FFN}_{\text{cheap}}$ outputs strictly zero at the start of training (regardless of the affine parameters in $\mathrm{LN}_{2,\text{cheap}}$); the cheap path therefore acts as an identity mapping within this residual sub-layer until the gate has learned meaningful routing decisions.

The controller first outputs a real-valued \emph{utility score} $u_i$ representing the unnormalized decision score for executing \texttt{full FFN}.
A soft gate
\begin{equation}
  p_i = \sigma(u_i / \tau)
\end{equation}
constructs a differentiable budget surrogate, where $\tau$ is the gate temperature.
The probability $p_i$ is used \emph{only} for budget/alive losses and does not directly determine the execution path.
The actual execution mask uses exact top-$k$: for a sequence of length~$T$, at each controlled layer we select
\begin{equation}
  k = \lceil \rho T \rceil
\end{equation}
tokens with the highest utility scores for \texttt{full FFN} execution; the remainder execute \texttt{cheap FFN}.
In this paper, $\rho = 0.5$.

\paragraph{Straight-through implementation detail.}
During training, the hard top-$k$ mask is $m_i^{\mathrm{hard}} \in \{0, 1\}$, where $m_i^{\mathrm{hard}} = 1$ if and only if token~$i$'s utility score $u_i$ ranks in the top~$k$.
The soft probability is $p_i = \sigma(u_i / \tau)$.
We construct a surrogate that equals the hard mask in the forward pass and routes gradients through $p_i$ in the backward pass:
\begin{equation}
  m_i^{\mathrm{st}} = \sg(m_i^{\mathrm{hard}} - p_i) + p_i,
\end{equation}
where $\sg(\cdot)$ denotes stop-gradient.
The actual mixture uses $m_i^{\mathrm{st}}$:
\begin{equation}
  \mathbf{h}_i = m_i^{\mathrm{st}} \cdot \mathbf{h}_i^{\text{full}} + (1 - m_i^{\mathrm{st}}) \cdot \mathbf{h}_i^{\text{cheap}}.
\end{equation}
The forward computation is therefore identical to hard top-$k$ routing; the backward pass propagates LM loss gradients to $u_i$ through the differentiable path $p_i = \sigma(u_i/\tau)$, which is the sole channel from LM loss to the gate.
Note that during training both \texttt{full} and \texttt{cheap} paths are computed for \emph{every} token (regardless of the top-$k$ selection), because the straight-through mixture and the JEPA target construction (\S\ref{sec:methods:gates}) both require access to both branch outputs.
At inference, only the hard top-$k$ rule is used and only the selected path is executed.

\subsection{Two Gate Training Methods}
\label{sec:methods:gates}

We compare two gate/controller training methods; all other settings are held constant.

\paragraph{MLP gate (G1, baseline).}
A two-layer MLP maps the current hidden state directly to a utility score:
\begin{equation}
  u_i = \mathbf{W}_2\,\mathrm{SiLU}(\mathbf{W}_1\,\mathbf{h}_i),
\end{equation}
where $\mathbf{W}_1 \in \mathbb{R}^{(d/4) \times d}$ and $\mathbf{W}_2 \in \mathbb{R}^{1 \times (d/4)}$, with hidden dimension $d_{\text{model}} / 4$.

\paragraph{JEPA-guided gate (G3).}
A shared predictor~$P$ is introduced.
The predictor is a two-layer MLP that takes the concatenation of a compressed context and an action embedding, producing low-dimensional outcome summaries for the \texttt{full} and \texttt{cheap} paths:
\begin{align}
  \mathbf{c}_i &= \mathbf{W}_c\,\sg(\mathbf{h}_i), \\
  q_a &= P([\mathbf{c}_i;\, \mathbf{e}_a]), \quad a \in \{\text{full}, \text{cheap}\},
\end{align}
where $P: \mathbb{R}^{d_c + d_a} \to \mathbb{R}^{d_s}$ consists of $\text{Linear} \to \text{SiLU} \to \text{Linear}$, $q_a \in \mathbb{R}^{d_s}$, and $\sg(\cdot)$ denotes stop-gradient so that the JEPA branch does not back-propagate into the backbone representation.
The decision head generates a utility score from
\begin{equation}
  u_i = D(q_{\text{full}},\; q_{\text{cheap}},\; q_{\text{full}} - q_{\text{cheap}}).
\end{equation}
With \texttt{future\_horizon} $= 1$, the predictor forecasts the outcome of executing \texttt{full} or \texttt{cheap FFN} at the \emph{current} controlled layer, not at subsequent layers.
The predictor's direct supervision comes from the JEPA alignment loss:
\begin{equation}
  \mathcal{L}_{\text{JEPA}} = \bigl(1 - \cos(q_{\text{full}}, z_{\text{full}})\bigr) + \bigl(1 - \cos(q_{\text{cheap}}, z_{\text{cheap}})\bigr),
\end{equation}
where $z_a = \sg\bigl(T(\mathbf{h}^{*}_{a,i})\bigr)$, $T: \mathbb{R}^d \to \mathbb{R}^{d_s}$ is a \emph{fixed} target projection head (orthogonally initialized, \texttt{requires\_grad=False}), and $\mathbf{h}^{*}_{a,i}$ is the hidden state obtained after token~$i$ actually executes path~$a$ at the current layer.
The target side does not participate in training; the predictor is trained to unilaterally align with the fixed target.
No EMA teacher is employed.

\begin{figure}[t]
  \centering
  \includegraphics[width=0.92\textwidth]{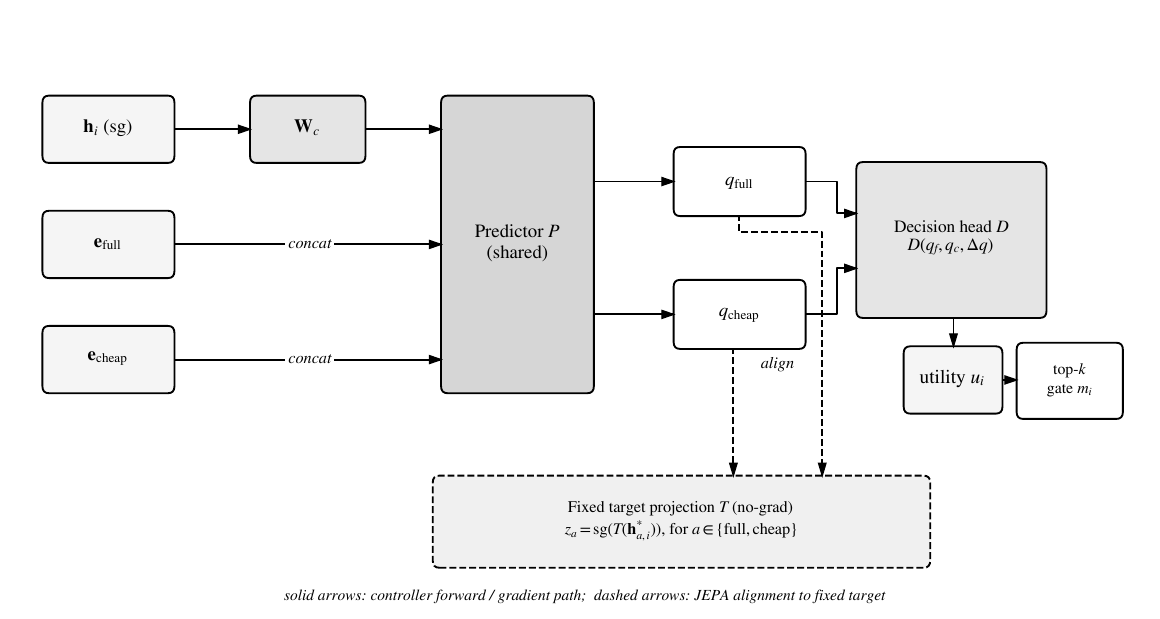}
  \caption{Architecture of the JEPA-guided gate (G3).
  Context projection $\mathbf{W}_c$ and the two discrete action embeddings $\mathbf{e}_{\text{full}}, \mathbf{e}_{\text{cheap}}$ feed the shared predictor $P$, yielding $q_{\text{full}}, q_{\text{cheap}}$.
  The decision head $D$ maps $(q_{\text{full}}, q_{\text{cheap}}, q_{\text{full}} - q_{\text{cheap}})$ to the scalar utility score $u_i$, which is then consumed by the top-$k$ gate/router to produce the execution mask $m_i$.
  The fixed target projection $T$ (\texttt{requires\_grad=False}) receives stop-gradient inputs from the post-execution hidden state $\mathbf{h}^{*}_{a,i}$ and produces $z_{\text{full}}, z_{\text{cheap}}$ for JEPA alignment; no EMA teacher is used.
  Solid arrows mark the controller forward/gradient path; dashed arrows mark the JEPA alignment to the fixed target.}
  \label{fig:g3_arch}
\end{figure}

\subsection{Online Utility/Rank Supervision}
\label{sec:methods:util}

In addition to JEPA alignment, training optionally employs a set of auxiliary losses based on online utility labels.
The motivation is that under frozen-backbone, controller-only training the gradient from the LM loss to the gate must traverse multiple layers, resulting in a weak and noisy signal; an explicit teacher signal acting directly on the gate output could, in principle, accelerate convergence.

\paragraph{Initial design rationale for the ``subsequent-all-full'' oracle.}
The specific label definition used below (``current layer forks into \texttt{full}/\texttt{cheap}; all subsequent controlled layers execute \texttt{full}'') was adopted as the \emph{first-version} oracle for three coupled reasons, not as a claim that it is the unique correct definition.
(i)~It is \emph{tractable} and deterministic: given the frozen backbone, the counterfactual trajectory for a single-layer decision can be computed by a fixed number of extra forward passes, independently of the current gate policy.
(ii)~It \emph{isolates} the marginal value of the \emph{current} routing decision by neutralising the confounding effect of subsequent-layer routing choices; this is the most natural way to obtain a token-level ``worth-computing'' label without introducing a second gate into the teacher.
(iii)~It \emph{avoids} making the teacher a moving target: its definition does not depend on the evolving gate policy, which simplifies both implementation and interpretation of the training signal.
Under these criteria the oracle is a reasonable initial choice; however, the same three properties also make the teacher structurally off-policy, as we now analyse, and this is the observation that motivates the retrospective diagnosis in~\S\ref{sec:discuss:retro}.

\paragraph{Utility label definition.}
For controlled layer~$l$, position~$t$, define
\begin{equation}
  u^{\star}_{l,t} = \widetilde{\mathrm{CE}}^{\text{cheap}}_{l,t} - \widetilde{\mathrm{CE}}^{\text{full}}_{l,t},
\end{equation}
where $\widetilde{\mathrm{CE}}$ is a backward-looking weighted-window cross-entropy:
\begin{equation}
  \widetilde{\mathrm{CE}}_{l,t} = \sum_{\delta=0}^{W-1} \gamma^{\delta}\,\mathrm{CE}_{l,t+\delta}.
\end{equation}
The label is computed by \emph{forking} the token's trajectory at layer~$l$---one branch executes \texttt{full}, the other executes \texttt{cheap} at the current layer---\emph{then forcing all subsequent controlled layers to execute \texttt{full}}, thereby measuring only the local marginal difference of this single-layer decision.
Thus $u^{\star} > 0$ indicates that executing \texttt{full} at this layer reduces the downstream window CE, while $u^{\star} \approx 0$ suggests that \texttt{cheap} suffices.

\begin{figure}[t]
  \centering
  \includegraphics[width=0.96\textwidth]{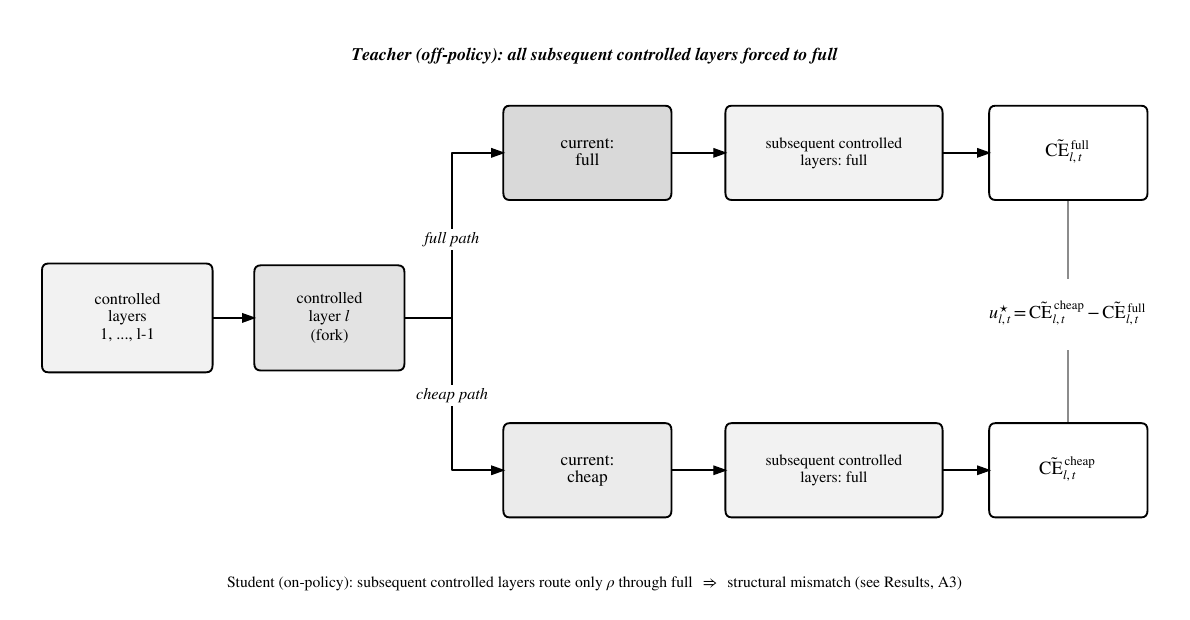}
  \caption{Counterfactual fork used to construct the utility label $u^{\star}$.
  At controlled layer $l$ and position $t$, the trajectory is forked: one branch executes \texttt{full} at the current layer, the other executes \texttt{cheap}.
  The teacher then \emph{forces} every subsequent controlled layer $l{+}1, \ldots, C$ to execute \texttt{full} on both branches, so the two branches produce windowed cross-entropies $\widetilde{\mathrm{CE}}^{\text{full}}_{l,t}$ and $\widetilde{\mathrm{CE}}^{\text{cheap}}_{l,t}$ whose difference defines $u^{\star}_{l,t}$.
  Under actual gated execution, only a fraction $\rho$ of subsequent tokens is routed through \texttt{full}; the teacher is therefore structurally off-policy with respect to the student, which is the source of the mismatch revisited in~\S\ref{sec:results:a3}.}
  \label{fig:util_label}
\end{figure}

\paragraph{Off-policy teacher bias.}
This label construction constitutes an \emph{off-policy teacher}: it assumes a counterfactual trajectory (``current layer forks; all subsequent layers execute full''), whereas training actually follows an on-policy execution (``current layer chosen by gate; subsequent layers also chosen by gate'').
Two structural characteristics of this bias deserve attention:

\begin{enumerate}[leftmargin=*,itemsep=2pt]
\item \textbf{The bias is directional, not random.}
Because the counterfactual trajectory assumes all subsequent layers execute \texttt{full}, any information loss from executing \texttt{cheap} at the current layer can be \emph{partially compensated} by subsequent full layers.
Under actual on-policy execution, $\rho = 0.5$ of subsequent tokens still execute cheap, making compensation weaker.
Consequently, $u^{\star}$ tends to \emph{systematically underestimate} the relative disadvantage of executing \texttt{cheap} under real conditions---a directional, non-random bias.

\item \textbf{Under controller-only training, this bias may accumulate over the course of training.}
The counterfactual world assumed by the teacher (``all subsequent full'') does not change with training, while the student (gate policy) evolves from initialization toward a non-trivial routing policy over 20k steps.
The distributional distance between teacher and student therefore \emph{tends to grow monotonically rather than fluctuate randomly}.
\end{enumerate}

We revisit this bias in light of the A3 ablation results in \S\ref{sec:results:a3}, and list on-policy utility (using the actual gate policy for trajectory rollout) as a necessary future experiment in \S\ref{sec:conclusion}.
The observation that utility/rank is harmful in \S\ref{sec:results:a3} is \textbf{strictly scoped to the current label definition} and cannot be directly extrapolated to all utility-style teachers.
Conceptually, the ``subsequent-all-full'' teacher / gated-execution student gap is a close analogue of the training--inference mismatch studied under scheduled sampling for sequence prediction~\citep{bengio2015scheduled} and the covariate-shift / compounding-error analysis of~\citet{ross2011dagger}; we do not claim that these references prove the A3 mechanism, but they provide useful conceptual support for treating the mismatch as a \emph{candidate} driver rather than an a~posteriori rationalisation.

\paragraph{Two auxiliary losses based on this label:}
\begin{align}
  \mathcal{L}_{\text{util}} &= \mathrm{Huber}(u_i,\, u^{\star}_{l,t}), \\
  \mathcal{L}_{\text{rank}} &= \mathbb{E}_{(i,j)}\bigl[\softplus\bigl(-\operatorname{sign}(u_i^{\star} - u_j^{\star})(u_i - u_j)\bigr)\bigr].
\end{align}
In implementation, to limit extra computation, online utility labels use a two-phase sparse strategy: (i)~during the first $S_{\text{warmup}}$ steps, no labels are computed and the gate trains only with LM + budget + alive; (ii)~thereafter, labels are recomputed every $I$ steps, with $\mathcal{L}_{\text{util}} = \mathcal{L}_{\text{rank}} = 0$ on non-refresh steps (stale labels are not cached or reused).

\subsection{Total Loss and Ablation Design}
\label{sec:methods:ablation}

The total training loss is:
\begin{equation}
  \mathcal{L} = \mathcal{L}_{\text{LM}} + \lambda_{\text{JEPA}}\mathcal{L}_{\text{JEPA}} + \lambda_{\text{util}}\mathcal{L}_{\text{util}} + \lambda_{\text{rank}}\mathcal{L}_{\text{rank}} + \lambda_b\mathcal{L}_{\text{budget}} + \lambda_a\mathcal{L}_{\text{alive}},
\end{equation}
where
\begin{align}
  \mathcal{L}_{\text{budget}} &= \frac{1}{C}\sum_{l=1}^{C}\bigl(\bar{p}_l - \rho\bigr)^2, \\
  \mathcal{L}_{\text{alive}} &= \frac{1}{C}\sum_{l=1}^{C}\relu\bigl(p_{\min} - \bar{p}_l\bigr),
\end{align}
and $\bar{p}_l$ is the batch-averaged gate probability at controlled layer~$l$.

All experiments use \textbf{controller-only} trainable scope: only the cheap path and control-related modules are trained---\texttt{cheap FFN}, the cheap-path-specific LayerNorm $\mathrm{LN}_{2,\text{cheap}}$, the decision head, and (for G3 only) the context projection $\mathbf{W}_c$, action embeddings $\mathbf{e}_a$, and predictor~$P$.
All other parameters---token/position embeddings, all normal blocks, the attention and \texttt{full FFN} in controlled layers, full-path LayerNorms, and the LM head---are frozen.
The \texttt{cheap FFN} is classified as ``control-related'' because it is an alternative branch activated only when the gate selects \texttt{cheap}, not the always-executed full-FFN backbone.
This design ensures that any observed differences can be attributed to the training of the gate, controller, and cheap path rather than to whole-model relearning.

To systematically disentangle the effects of each auxiliary loss, we design four ablation groups (all starting from G3):

\begin{itemize}[leftmargin=*,itemsep=2pt]
\item \textbf{A1 (jepa-zero-arch):} Retain the G3 architecture but set $\lambda_{\text{JEPA}} = 0$.
The predictor remains in the forward pass but receives no direct alignment constraint, updating only through indirect gradients from util/rank/LM.

\item \textbf{A2 (shuffled-target):} Retain the JEPA loss but replace each token's real \texttt{full}/\texttt{cheap} target projection with a derangement (fixed-point-free permutation) across the batch$\,\times\,$sequence dimension.
The predictor still converges toward a real target projection, but one that does not correspond to the current token's actual execution outcome.

\item \textbf{A3 (no-util-loss):} Set $\lambda_{\text{util}} = \lambda_{\text{rank}} = 0$, applied to both the G3 architecture (A3-G3) and the G1 architecture (A3-G1).
The budget executor is unaffected; the top-$k$ rule is retained.

\item \textbf{A4 (weight-sweep):} Fix all other settings and set $\lambda_{\text{JEPA}}$ to 0.25 and 2.0 (default 1.0).
\end{itemize}

\section{Experimental Setup}
\label{sec:setup}

\paragraph{Model.}
Decoder-only Transformer, 12~layers, hidden dimension 640, 10 attention heads, FFN hidden dimension 2560; vocabulary size 151{,}665 (Qwen2.5-0.5B tokenizer).
\textbf{Total parameters $\approx\,157.5$M} (G1: 157{,}469{,}124; G3: 157{,}530{,}948; the difference arises from the predictor and action embeddings).
\textbf{Trainable controller parameters:} 828{,}484 (MLP gate) / 849{,}348 (JEPA-guided gate); all other parameters are frozen.
The last 4 layers are controlled layers, with cheap-path rank $r = 80$, predictor context dimension $d_c = 128$, summary dimension $d_s = 64$, and action embedding dimension $d_a = 16$.

\paragraph{Data.}
We use the \texttt{sample-10BT} subset of \texttt{HuggingFaceFW/fineweb-edu}~\citep{penedo2024fineweb}.
For reproducibility, we extract the first 300{,}000 non-empty \texttt{text} samples in their original streaming order, write them as a fixed local JSONL subset, then train; 290{,}000 samples serve as the training set and 10{,}000 as the validation set.
Sequence length is 256, tokenized with the Qwen2.5-0.5B tokenizer~\citep{qwen2024qwen25} (vocabulary size 151{,}665).
No cross-corpus contamination check, cross-domain evaluation, or larger-scale data validation has been performed; all conclusions hold only on this fixed subset.

\paragraph{Training.}
Controller-only training (the precise trainable scope is defined in \S\ref{sec:methods:ablation}), initialized from a shared pretrained checkpoint from an earlier full-only training phase.
Optimizer: AdamW~\citep{loshchilov2019adamw}; learning rate $2 \times 10^{-4}$; weight decay 0.01; gradient clipping threshold 1.0; linear warmup for 1{,}000 steps followed by cosine decay to $\mathrm{lr} \times 0.01$; bf16 mixed precision.
Batch size 16; 20{,}000 training steps; validation every 500 steps (up to 16 batches, 256 samples per evaluation).
Budget parameter $\rho = 0.5$; gate temperature $\tau = 2.0$; minimum alive ratio $p_{\min} = 0.05$.
\textbf{Standard recipe loss weights:} $\lambda_{\text{JEPA}} = 1.0$, $\lambda_{\text{util}} = 1.0$, $\lambda_{\text{rank}} = 0.2$, $\lambda_b = 1.0$, $\lambda_a = 0.5$.
Utility label: window $W = 8$, decay $\gamma = 0.5$, warmup $S_{\text{warmup}} = 100$, refresh interval $I = 5$.

\paragraph{Parameter count.}
MLP gate trainable parameters: 828{,}484; JEPA-guided gate: 849{,}348 (difference $\approx 2.5\%$, from predictor and action embeddings).
To rule out parameter count as a confound, we include a \texttt{G1-costmatch} control with an enlarged MLP gate hidden dimension (trainable parameters: 902{,}956, comparable to or slightly exceeding G3).

\paragraph{Metrics.}
We use \textbf{dynamic/cumulative metrics} as primary references and \textbf{single-point metrics} as supplements, reducing the risk of cherry-picking the luckiest checkpoint:

\begin{itemize}[leftmargin=*,itemsep=1pt]
\item \emph{Primary (dynamic/cumulative):} validation \texttt{eval lm\_loss} averaged over the 0--10k and 0--20k step windows (\texttt{avg\_lm\_0\_10k}, \texttt{avg\_lm\_0\_20k}); first step to hit fixed thresholds (5.075 / 5.070 / 5.065); training \texttt{grad\_norm} mean (used only jointly with diagnostic quantities).
\item \emph{Supplementary (single-point):} 20k-step endpoint \texttt{eval lm\_loss}; minimum \texttt{eval lm\_loss} over the entire run (\texttt{best\_eval\_lm}).
\item \emph{Diagnostic:} gate activation probability; actual full ratio; predictor output $\ell_2$ distance and cosine similarity (\texttt{diag\_qf\_qc\_l2}, \texttt{diag\_qf\_qc\_cos}); utility score variance (\texttt{diag\_util\_score\_var}).
\end{itemize}

\paragraph{0.005 heuristic reference threshold.}
The 0.005 threshold is defined \textbf{only on the 20k-step endpoint \texttt{eval lm\_loss}} single-point metric: two configurations are termed ``within the heuristic reference threshold'' if and only if their endpoint \texttt{eval lm\_loss} difference satisfies $|\Delta| \le 0.005$.
This value was set before observing experimental results, based on the inter-seed fluctuation magnitude in pilot training (one short run each of G1 and G3); it is a heuristic rather than a statistically derived non-inferiority bound.
It is \emph{not} applied to \texttt{best\_eval\_lm}, \texttt{avg\_lm}, \texttt{hit\_step}, or any non-LM metric.
When per-seed $|\Delta|$ values for \texttt{best\_eval\_lm} are compared against 0.005 (as in \S\ref{sec:results:a3}), this should be understood as a rough order-of-magnitude reference, not a formal non-inferiority test.

\paragraph{Threshold provenance.}
The thresholds 5.075 / 5.070 / 5.065 were selected based on pilot training (one short run each of G1 and G3) prior to the main experiment, designed to span ``easy / medium / hard to hit'' tiers within the observed eval LM range.
They were fixed before the main experiment summary data were generated.

\paragraph{Cross-experiment comparison metric.}
We uniformly use \texttt{eval lm\_loss}; the mixed \texttt{total\_loss} is not used because auxiliary loss weights differ across experiments.

\paragraph{Compute proxy definition.}
All ``training-time compute'' and ``inference-time compute'' numbers in this paper are \emph{relative cost proxies} in units of FFN-equivalent token-layers, \textbf{not wall-clock time}.
The full-path cost per FFN execution unit is~1; the cheap-path cost is $r / d_{\text{ff}} \approx 0.031$.
Controller, predictor, and target projection overhead are included in training cost by FLOPs equivalence; oracle utility label construction (every $I{=}5$ steps) is similarly included.
For calibration, we additionally report \emph{measured} wall-clock for the full 20k-step runs used in~\S\ref{sec:results:a3}: on a single NVIDIA V100 (32\,GB) with 32\,GB system memory, G3 takes ${\approx}\,2.87\mathrm{h}$ and A3-G3 takes ${\approx}\,1.75\mathrm{h}$ (${\approx}\,39\%$ reduction).
The proxy and the measured wall-clock are directionally consistent but not numerically identical (proxy: $30.0\%$; measured: ${\approx}\,39\%$), because the proxy does not model kernel-launch overheads, data-loading costs, or the \texttt{torch.no\_grad()} forwards performed during online label construction.
We therefore report the FLOPs proxy and the wall-clock number side-by-side rather than conflating them into a single efficiency claim.

\paragraph{Random seeds (per-group declaration).}
Main comparison (G1-base vs.\ G3), ablations A1 / A2 / A4 / A3-G3 / A3-G1: \textbf{3 seeds (42, 123, 7)}.
Supplementary control \texttt{G1-costmatch}: \textbf{1 seed (42) only}; conclusions involving this group are marked ``single-seed preliminary observation.''
25\% budget supplementary experiment (\S\ref{sec:results:25pct}): \textbf{2 seeds (42, 123)}; evidence level explicitly lower than the main experiment.
This section is the \textbf{sole authoritative statement} on seed coverage; any simplified wording elsewhere defers to this section.

\section{Results}
\label{sec:results}

\begin{figure}[t]
  \centering
  \includegraphics[width=\textwidth]{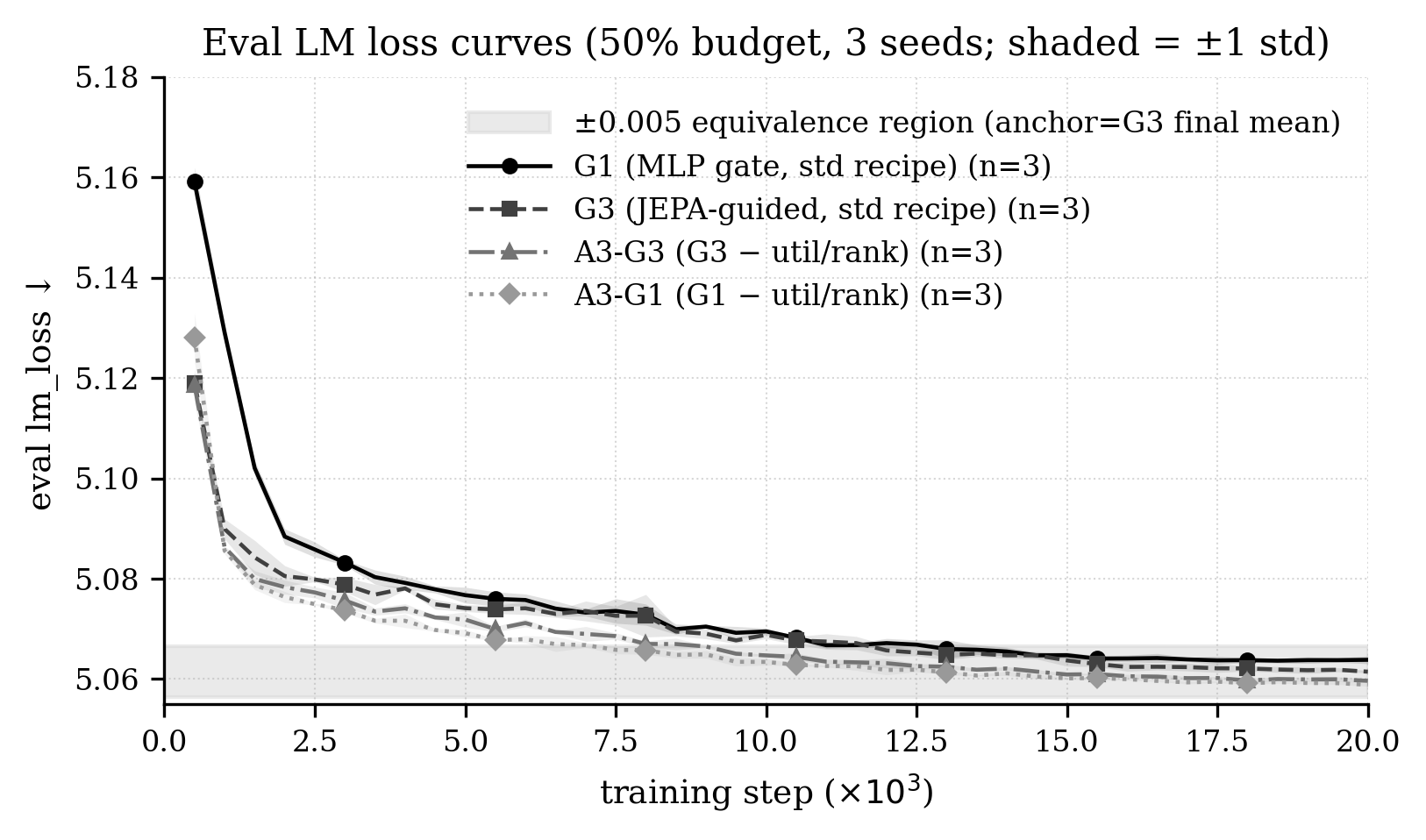}
  \caption{Validation \texttt{eval lm\_loss} learning curves for four core configurations at 50\% budget (3 seeds; solid line = mean, shaded region = $\pm 1$ std).
  The gray horizontal band marks $\pm 0.005$ around the G3 endpoint mean as a heuristic reference (for \emph{endpoint} \texttt{eval lm\_loss} only; not a formal non-inferiority test).
  Key reading points:
  (a)~G3 consistently outperforms G1 in the early-to-mid training phase under the standard recipe (\S\ref{sec:results:main});
  (b)~at the endpoint, G1 enters the upper edge of the equivalence band---the G3 vs.\ G1 endpoint LM difference falls within the 0.005 margin;
  (c)~A3-G3 and A3-G1 nearly overlap after ${\approx}\,7$k steps, with endpoints at the lower edge of the band---their LM difference is not meaningfully detectable (\S\ref{sec:results:a3});
  (d)~both A3 curves consistently outperform the standard-recipe G3 curve throughout the main training phase.}
  \label{fig:eval_curves}
\end{figure}

\subsection{G1 vs.\ G3 Under the Standard Recipe}
\label{sec:results:main}

Table~\ref{tab:main} reports the three-seed mean comparison under the standard recipe (all auxiliary losses active).
Under identical 50\% budget and identical other settings, the JEPA-guided gate (G3) achieves better best / avg / hit metrics than the MLP gate (G1) across 3/3 seeds, with consistent direction.
The training-time \texttt{grad\_norm} mean decreases substantially (${\approx}\,10.3\times$; interpretation requires joint reading with diagnostics in \S\ref{sec:results:a1} and \S\ref{sec:discuss:grad}).
The 20k-step endpoint LM falls within the preset 0.005 heuristic reference threshold ($\Delta = -0.0021$).

\begin{table}[t]
\centering
\caption{Main comparison under the standard recipe: G1 vs.\ G3 (50\% budget, 3 seeds, mean $\pm$ std). ``best lm'' omits the leading ``5.'' (\eg, $.0636 = 5.0636$); ``avg 10k''/``avg 20k'' are mean eval lm over steps 0--10k / 0--20k; ``hit $X$'' is the first step at which eval lm crosses threshold $5.0X$; ``grad'' is the training-time \texttt{grad\_norm} mean. All metrics are ``lower is better.''}
\label{tab:main}
\small
\setlength{\tabcolsep}{3.5pt}
\begin{tabular}{@{}lccccccc@{}}
\toprule
Config & best lm & avg 10k & avg 20k & hit 75 & hit 70 & hit 65 & grad \\
\midrule
G1-base & $.0636{\scriptstyle \pm .0005}$ & $.0843{\scriptstyle \pm .0005}$ & $.0747{\scriptstyle \pm .0005}$ & $6333{\scriptstyle \pm 289}$ & $9167{\scriptstyle \pm 577}$ & $14167{\scriptstyle \pm 1258}$ & $83.33{\scriptstyle \pm 12.29}$ \\
G3 & $.0614{\scriptstyle \pm .0002}$ & $.0776{\scriptstyle \pm .0005}$ & $.0707{\scriptstyle \pm .0005}$ & $4833{\scriptstyle \pm 289}$ & $8333{\scriptstyle \pm 289}$ & $12667{\scriptstyle \pm 764}$ & $8.06{\scriptstyle \pm 0.23}$ \\
\bottomrule
\end{tabular}
\end{table}

\paragraph{Ruling out parameter count as a confound (single-seed preliminary observation).}
\texttt{G1-costmatch} (enlarged MLP gate hidden dimension; trainable parameters 902{,}956, comparable to G3's 849{,}348) has been run for \textbf{seed 42 only}.
Under this seed, \texttt{G1-costmatch} achieves best eval lm $= 5.0628$, between same-seed G1 ($5.0642$) and G3 ($5.0613$), \emph{not matching G3}.
This is directional evidence that the G3 improvement cannot be fully explained by controller parameter count, but with only 1~seed, the evidence level is lower than the main comparison.

Thus, under the standard recipe, the JEPA-guided gate improves gate/controller optimization dynamics relative to the MLP gate, with endpoint LM within the heuristic reference threshold.
Section~\ref{sec:results:a3} will revisit this comparison after removing util/rank.

\subsection{A1: The Predictor Requires Direct Supervision to Avoid Representation Collapse}
\label{sec:results:a1}

Removing JEPA direct supervision (A1) yields consistent degradation across 3 seeds.

\begin{table}[t]
\centering
\caption{A1 ablation and main comparison (3 seeds, mean $\pm$ std). ``best lm'' omits the leading ``5.''; ``$\ell_2$'' and ``cos'' are \texttt{diag\_qf\_qc\_l2} / \texttt{diag\_qf\_qc\_cos} (predictor collapse diagnostics); ``---'' = not applicable.}
\label{tab:a1}
\small
\setlength{\tabcolsep}{3.5pt}
\begin{tabular}{@{}lcccccccc@{}}
\toprule
Config & best lm & avg 10k & avg 20k & hit 75 & hit 70 & grad & $\ell_2$ & cos \\
\midrule
G3 & $.0614{\scriptstyle \pm .0002}$ & $.0776{\scriptstyle \pm .0005}$ & $.0707{\scriptstyle \pm .0005}$ & 4833 & 8333 & 8.06 & --- & --- \\
A1 & $.0735{\scriptstyle \pm .0019}$ & $.0848{\scriptstyle \pm .0007}$ & $.0808{\scriptstyle \pm .0004}$ & 18333 & --- & 2.27 & 0.005 & 1.001 \\
\bottomrule
\end{tabular}
\end{table}

A1's best eval lm degrades by $+0.0121$ relative to G3; avg lm (0--10k) degrades by $+0.0072$; the first hit of 5.075 shifts from 4{,}833 to 18{,}333 steps, and no seed hits 5.070 within 20k steps.

The critical diagnostic: A1's final \texttt{diag\_qf\_qc\_l2} is merely $0.004$--$0.008$, while A2 (which retains JEPA loss) shows $0.82$--$0.90$.
That is, \textbf{without JEPA alignment, the predictor outputs for \texttt{full} and \texttt{cheap} nearly collapse}---the ``two-action comparison'' feature on which the gate depends is effectively constant.
Simultaneously, \texttt{diag\_util\_score\_var} drops to the $10^{-5}$ order, indicating that utility scores themselves are near-constant; top-$k$ ranking under these conditions is dominated by random noise.

\paragraph{A note on diagnostic choice.}
We also monitor \texttt{diag\_qf\_qc\_cos} but find it $\approx 1.0$ across \emph{all} runs (including non-collapsed ones), with the collapsed A1 runs recording $\approx 1.002$; under bf16, a cosine reading of $1.0 \pm 0.002$ lies within the expected numerical-artefact range and we do not interpret this as a physically meaningful super-unity similarity.
In the current architecture, the action information encoded by the predictor resides primarily in the \emph{magnitude and small orthogonal components} of $q_f, q_c$, with directions nearly collinear.
\textbf{Subsequent work on action-conditional predictors should use $\ell_2$ distance, not cosine similarity, as the primary collapse diagnostic}---a minor but potentially useful observation.

\paragraph{Interpretation.}
A1 demonstrates that the predictor cannot acquire useful representations through indirect gradients alone under controller-only training; direct JEPA alignment is a necessary condition for this architecture to function.
However, A1 alone does not prove that the representation provided by JEPA is the main source of LM improvement---this requires joint consideration with A2 and A3.

Note: A1's \texttt{grad\_norm} is paradoxically the \emph{lowest} ($2.27$), but this reflects parameter-space degeneracy from predictor collapse (near-zero score variance) rather than healthier optimization.
Gradient norms should always be read jointly with LM metrics and predictor diagnostics.

\subsection{A2: Disrupting Token-Wise Pairing Causes Mild but Consistent Degradation}
\label{sec:results:a2}

With JEPA loss retained but target projections randomly deranged (A2), degradation is much smaller than A1 but directionally consistent across 3/3 seeds.

\begin{table}[t]
\centering
\caption{A2 ablation (3 seeds, mean $\pm$ std). Column shorthand follows Table~\ref{tab:main}; ``$\ell_2$'' is \texttt{diag\_qf\_qc\_l2}. Hit~75 is omitted: per-seed values for A2 and G3 differ by $\leq$500 steps, making the column uninformative for this comparison; hit~70 and hit~65 show the meaningful separation. All metrics are ``lower is better.''}
\label{tab:a2}
\small
\setlength{\tabcolsep}{3.5pt}
\begin{tabular}{@{}lcccccc@{}}
\toprule
Config & best lm & avg 10k & avg 20k & hit 70 & hit 65 & $\ell_2$ \\
\midrule
G3 & $.0614{\scriptstyle \pm .0002}$ & $.0776{\scriptstyle \pm .0005}$ & $.0707{\scriptstyle \pm .0005}$ & 8333 & 12667 & --- \\
A2 & $.0645{\scriptstyle \pm .0000}$ & $.0785{\scriptstyle \pm .0001}$ & $.0724{\scriptstyle \pm .0002}$ & 9667 & 16167 & 0.850 \\
\bottomrule
\end{tabular}
\end{table}

A2 vs.\ G3: best eval lm degrades by $+0.0030$; avg lm (0--10k) is nearly flat; stricter thresholds 5.070/5.065 are delayed to 9{,}667/16{,}167 steps.

Diagnostics confirm that A2 does \emph{not} exhibit representation collapse: \texttt{diag\_qf\_qc\_l2} $= 0.82$--$0.90$, far above A1's near-zero level.
This is consistent with the predictor still converging toward ``some real target projection''---just one that does not correspond to the current token's own execution outcome.

\paragraph{Interpretation.}
Two observations merit separate consideration.
First, A2 does not collapse, indicating that a shuffled target still provides a sufficiently ``structured'' auxiliary objective to maintain non-trivial predictor outputs---in stark contrast to A1.
Second, the mild degradation relative to G3 shows that strict token-wise pairing does provide measurable benefit under the full recipe (with util/rank), but its magnitude is smaller than the choice of ``whether JEPA alignment exists at all'' (A1 vs.\ G3).
Importantly, A2 was conducted \emph{with utility/rank retained}, so it cannot cleanly separate (a)~the intrinsic benefit of strict semantic pairing from (b)~utility/rank's partial compensation for misaligned targets.
Full separation requires an A5 cross-ablation (shuffled-target $+$ no-util/rank), listed as future work in \S\ref{sec:conclusion}.

\subsection{A3: Removing Util/Rank Improves All Primary Metrics}
\label{sec:results:a3}

This finding directly contradicts the intuition that ``stacking more auxiliary supervision is generally better,'' and also revises the G1 vs.\ G3 comparison from \S\ref{sec:results:main}: without util/rank, the two gates become indistinguishable in LM.

\begin{table}[t]
\centering
\caption{A3 ablation (3 seeds, mean $\pm$ std, 50\% budget). Column shorthand follows Table~\ref{tab:main}; ``comp.''\ is the training-time FFN-equivalent FLOPs proxy \texttt{compute\_vs\_full}; ``wall (h)'' is measured per 20k-step run. All metrics are ``lower is better.''}
\label{tab:a3}
\small
\setlength{\tabcolsep}{3pt}
\begin{tabular}{@{}lcccccccc@{}}
\toprule
Config & best lm & avg 10k & avg 20k & hit 70 & hit 65 & grad & comp. & wall (h) \\
\midrule
G1-base & $.0636{\scriptstyle \pm .0005}$ & $.0843{\scriptstyle \pm .0005}$ & $.0747{\scriptstyle \pm .0005}$ & 9167 & 14167 & 83.33 & 1.513 & --- \\
G3 & $.0614{\scriptstyle \pm .0002}$ & $.0776{\scriptstyle \pm .0005}$ & $.0707{\scriptstyle \pm .0005}$ & 8333 & 12667 & 8.06 & 1.527 & 2.87 \\
A3-G3 & $.0597{\scriptstyle \pm .0004}$ & $.0743{\scriptstyle \pm .0004}$ & $.0678{\scriptstyle \pm .0004}$ & 6167 & 9833 & 0.51 & 1.068 & 1.75 \\
A3-G1 & $.0589{\scriptstyle \pm .0003}$ & $.0729{\scriptstyle \pm .0002}$ & $.0667{\scriptstyle \pm .0002}$ & 4667 & 8333 & 0.48 & 1.054 & --- \\
\bottomrule
\end{tabular}
\end{table}

The data show:

\paragraph{(i) Removing util/rank consistently improves LM for both architectures.}
A3-G3 vs.\ G3: best eval lm improves in 3/3 seeds (paired mean $\Delta \approx -0.0018$); avg lm and all three threshold hit steps likewise improve.
A3-G1 vs.\ G1 shows the same direction with a larger effect (paired mean $\Delta \approx -0.0047$).
Both paired mean deltas fall within $|\Delta| < 0.005$, so under the strict \S\ref{sec:setup} definition of the 0.005 margin (applicable only to endpoint \texttt{eval lm\_loss}), we do \emph{not} claim A3's LM improvement as ``detectable past the 0.005 threshold.''
Instead, we characterize it as \textbf{directionally stable}: the improvement is consistent across all metrics $\times$ all 3 seeds.

\paragraph{(ii) Without util/rank, simple MLP gate and JEPA-guided gate are indistinguishable in LM.}
A3-G1 vs.\ A3-G3 (best eval lm, negative~$=$~A3-G1 better): per-seed $\Delta$ values are $-0.0008$, $-0.0004$, and $-0.0012$; direction is consistent in 3/3 seeds toward A3-G1, but all $|\Delta| < 0.005$.
The strictly claimable statement is:

\begin{quote}
``In 3/3 seeds, the A3-G1 vs.\ A3-G3 best eval lm per-seed $|\Delta| \in [0.0004, 0.0012]$, all $< 0.005$.
Under the current recipe, the JEPA predictive auxiliary does not provide detectable additional LM benefit.
(This claim is strictly scoped to the A3 recipe without util/rank; the JEPA value under the standard recipe is given in \S\ref{sec:results:main}.)''
\end{quote}

\paragraph{(iii) Score calibration behavior differs between architectures, but routing outcomes are identical.}
A3-G3's \texttt{mean\_gate\_prob} stabilizes at $\approx 0.496$ across seeds; A3-G1 shows marked divergence ($0.622 / 0.372 / 0.370$ for seeds 42 / 123 / 7).
Despite the $\approx 0.25$ inter-seed swing in absolute score scale, \textbf{all seeds maintain \texttt{full\_ratio} precisely at 0.5} due to the hard top-$k$ constraint.
This confirms that the absolute score scale has no effect on routing outcomes---only relative ordering matters---consistent with working hypothesis (H2) below.

\paragraph{(iv) Training-time overhead drops, but the claim is proxy-first.}
A3 skips oracle utility label construction.
The training-time \texttt{compute\_vs\_full} of G3 drops from $1.527$ to $1.068$ for A3-G3 and to $1.054$ for A3-G1, while inference-time \texttt{infer\_vs\_full} is unchanged ($\approx 0.547$--$0.553$).
In our setup, the corresponding \emph{measured} wall-clock for a full 20k-step run drops from ${\approx}\,2.87\mathrm{h}$ (G3) to ${\approx}\,1.75\mathrm{h}$ (A3-G3), or ${\approx}\,39\%$.
The FLOPs proxy and wall-clock are directionally consistent but not numerically identical: the proxy predicts a $30.0\%$ reduction whereas the measured reduction is $39\%$, reflecting implementation-specific overheads (kernel-launch costs, online-label forward/backward bookkeeping) that the proxy does not model.
The correct reading is therefore: under this regime, \textbf{removing util/rank improves LM while directionally reducing training-time cost}; we report both the FLOPs proxy and the wall-clock number rather than conflating them into a single engineering claim.

\paragraph{Interpretation.}
The strictly claimable statement is the phenomenon itself:

\begin{quote}
``\textbf{Under the current standard recipe} ($\lambda_{\text{util}} = 1.0$, $\lambda_{\text{rank}} = 0.2$, ``subsequent-all-full'' oracle label definition), simultaneously introducing utility regression and pairwise rank constitutes a \textbf{net-negative contribution to LM} in 3/3 seeds, and the JEPA architecture provides no detectable LM benefit under this recipe.''
\end{quote}

Regarding \emph{why} net-negative, we propose two mutually compatible \textbf{working hypotheses}---noting that our experimental matrix \emph{cannot directly distinguish them from other candidates}:

\begin{enumerate}[label=(H\arabic*),leftmargin=*]
\item \textbf{Off-policy teacher--on-policy execution distribution mismatch.}
As discussed in \S\ref{sec:methods:util}, the utility label's counterfactual trajectory assumes ``all subsequent full,'' and this bias may accumulate over training under controller-only settings.
$\lambda_{\text{util}} = 1.0$ (same magnitude as LM loss) may push scores toward a structurally biased target, limiting the LM loss's ability to freely shape scores.

\item \textbf{Top-$k$ routing is insensitive to absolute score scale.}
Routing depends only on top-$k$; absolute score values do not affect execution---consistent with the observation in~(iii).
Utility/rank anchors the absolute scale and pairwise absolute differences to $u^{\star}$'s numerical scale, which is not what top-$k$ routing requires.
\end{enumerate}

These are \textbf{not the only candidate explanations}.
At least the following alternatives \emph{cannot be ruled out} with the current data:

\begin{enumerate}[label=(H\arabic*),start=3,leftmargin=*]
\item The current weight combination $(\lambda_{\text{util}}, \lambda_{\text{rank}}) = (1.0, 0.2)$ itself lies in an ``excessively strong util/rank'' region.
\item The Huber term and LM term exhibit local gradient-direction conflict at the token level (a mechanism separate from teacher--student distribution mismatch).
\item The oracle label definition (``subsequent all full'') itself introduces excessive bias (\S\ref{sec:methods:util}).
\end{enumerate}

Distinguishing (H1)--(H5) requires three types of experiments absent from this paper: (a)~$\lambda_{\text{util}}$ sweep; (b)~on-policy utility; (c)~separate util-only vs.\ rank-only ablation.
All are listed as future work in \S\ref{sec:conclusion}.

A retrospective diagnosis of why the ``subsequent-all-full'' oracle is a reasonable initial design yet structurally off-policy is deferred to \S\ref{sec:discuss:retro}.

\subsection{A4: Prediction Supervision Weight Is Insensitive}
\label{sec:results:a4}

Adjusting $\lambda_{\text{JEPA}}$ from 1.0 to 0.25 or 2.0, the three-seed aggregate results are nearly unchanged:

\begin{table}[t]
\centering
\caption{$\lambda_{\text{JEPA}}$ sweep (3 seeds, mean $\pm$ std). Column shorthand follows Table~\ref{tab:main}. All metrics are ``lower is better.''}
\label{tab:a4}
\small
\setlength{\tabcolsep}{4pt}
\begin{tabular}{@{}lccc@{}}
\toprule
$\lambda_{\text{JEPA}}$ & best lm & avg 10k & avg 20k \\
\midrule
0.25 & $.0606{\scriptstyle \pm .0005}$ & $.0787{\scriptstyle \pm .0006}$ & $.0709{\scriptstyle \pm .0005}$ \\
1.0 (= G3) & $.0614{\scriptstyle \pm .0002}$ & $.0776{\scriptstyle \pm .0005}$ & $.0707{\scriptstyle \pm .0005}$ \\
2.0 & $.0617{\scriptstyle \pm .0004}$ & $.0775{\scriptstyle \pm .0006}$ & $.0708{\scriptstyle \pm .0005}$ \\
\bottomrule
\end{tabular}
\end{table}

Differences across all three settings are within same-seed standard deviations, in stark contrast to A1's marked degradation.
This indicates that, under the current architecture and recipe, \textbf{what matters is whether the JEPA alignment constraint is present, not its precise weight}: once imposed, the predictive supervision constrains the predictor to a non-degenerate, full/cheap-discriminable low-dimensional space, enabling the decision head to produce informative scores.
This is consistent with the collapse diagnostic from A1.
Note that this sweep was conducted under the standard recipe (with util/rank); whether the same insensitivity holds in the A3 recipe (without util/rank) has not been tested.

\subsection{25\% Budget Preliminary Supplement (2 Seeds, Lower Evidence Level)}
\label{sec:results:25pct}

To check whether the optimization dynamics difference from \S\ref{sec:results:main} appears only at 50\% budget, we run G1-base $\times$ 2 seeds and G3 $\times$ 2 seeds at 25\% budget (each controlled layer routes only 25\% of tokens to full).
\textbf{With only 2 seeds, this section's conclusions carry lower evidence weight than the main experiment.}

\begin{table}[t]
\centering
\caption{25\% budget supplement (2 seeds). Column shorthand follows Table~\ref{tab:main}. All metrics are ``lower is better.''}
\label{tab:25pct}
\small
\setlength{\tabcolsep}{3.5pt}
\begin{tabular}{@{}lcccccc@{}}
\toprule
Config & best lm & avg 10k & avg 20k & hit 75 & hit 70 & grad \\
\midrule
G1-base & $.0599{\scriptstyle \pm .0000}$ & $.0889{\scriptstyle \pm .0010}$ & $.0753{\scriptstyle \pm .0006}$ & 4500 & 7500 & 138.65 \\
G3 & $.0604{\scriptstyle \pm .0003}$ & $.0790{\scriptstyle \pm .0005}$ & $.0707{\scriptstyle \pm .0004}$ & 3500 & 7750 & 47.80 \\
\bottomrule
\end{tabular}
\end{table}

At 25\% budget, the G3 vs.\ G1 \texttt{best eval lm} per-seed differences are $\{+0.0003, +0.0007\}$ (sign: G3 slightly \emph{worse}), both $\ll 0.005$; the corresponding endpoint \texttt{eval lm\_loss} per-seed differences are $\{+0.0003, +0.0008\}$, also well within the 0.005 heuristic reference threshold.
G3 shows better avg lm (0--10k) by $\approx -0.010$ and earlier hit at the looser threshold 5.075 (3{,}500 vs.\ 4{,}500), but does \emph{not} lead at the stricter threshold 5.070 ($7{,}750 \pm 354$ vs.\ $7{,}500 \pm 0$).
We interpret the 25\% budget results as: \textbf{G3 still shows better early-to-mid optimization trend and lower \texttt{grad\_norm} at a tighter budget, but ``earlier threshold hit'' holds only for looser thresholds, not universally}.

\paragraph{Important caveat.}
Combining this section with \S\ref{sec:results:main}: the G3 vs.\ G1 endpoint \texttt{eval lm\_loss} difference is not monotonic across $\rho$.
At $\rho = 0.5$, per-seed differences favor G3; at $\rho = 0.25$, per-seed signs reverse---but both fall within the 0.005 heuristic reference threshold.
The more accurate reading is \textbf{``within the heuristic reference threshold at both budgets,''} rather than ``G3 monotonically improves G1 in endpoint LM.''
The A3 ablation at 25\% budget has not been completed; whether the A3 phenomenon replicates at tighter budgets is an open question.

\section{Discussion}
\label{sec:discuss}

\subsection{Two Distinct Roles of Auxiliary Losses}
\label{sec:discuss:roles}

Synthesizing the evidence from \S\ref{sec:results}, the auxiliary losses serve two \textbf{separable, non-interchangeable} roles in gate training:

\begin{itemize}[leftmargin=*,itemsep=2pt]
\item \textbf{Feature shaping} (borne by the JEPA loss).
The JEPA alignment constraint enforces that the two action-conditional outputs of the predictor, $q_{\text{full}}$ and $q_{\text{cheap}}$, remain non-degenerate.
The target $z_a$ is constructed from the current token's \emph{within-layer} execution outcome---a \textbf{local, current-layer, action-conditional branch target}.
While not strictly on-policy in the RL sense, it does \emph{not} involve the long-horizon counterfactual rollout used by the utility label, and therefore introduces less distributional shift.
A1 proves this role cannot be replaced by indirect gradients; A2 shows that strict token-wise pairing strengthens it but is not strictly necessary; A4 shows that once the constraint exists, its weight is insensitive in the 0.25--2.0 range.

\item \textbf{Score anchoring} (borne by util/rank).
Through an oracle-style, explicit long-horizon counterfactual teacher, the utility/rank losses anchor the \emph{numerical scale} of the gate output to the weighted-window CE difference.
As established in \S\ref{sec:results:a3}, this role constitutes a net-negative contribution to LM under the current recipe.
Top-$k$ routing requires only relative ordering, not numerical scale; and the teacher--execution distribution mismatch (\S\ref{sec:methods:util}) may actively harm LM optimization.
\end{itemize}

\paragraph{Convergent picture from A1 and A3.}
Section~\ref{sec:results:main} establishes that G3 improves the optimisation dynamics of G1 under the standard recipe, and \S\ref{sec:results:a3} establishes that once util/rank is removed, the LM difference between G3 and G1 falls within the heuristic reference threshold.
Both facts are supported by data and mutually compatible.
Together, they yield a more precise description:

\begin{quote}
``\textbf{JEPA alignment guarantees that the two action-conditional outputs of the predictor remain non-collapsed and discriminable} (A1 provides the counter-evidence; A4 shows that this property is maintained across a wide $\lambda_{\text{JEPA}}$ range).
\textbf{Whether this discriminability translates into detectable LM benefit depends on whether util/rank is simultaneously present in the recipe}: when present, it appears as improved early-to-mid optimisation dynamics (\S\ref{sec:results:main}); when absent, only differences on the order of the 0.005 margin remain (\S\ref{sec:results:a3}).''
\end{quote}

A natural working hypothesis is that JEPA alignment acts as a \emph{buffer} for the biased oracle teacher---the LM value of JEPA partly derives from mitigating the off-policy bias of util/rank rather than from independently providing new training signal.
This requires two independent experiments to verify: (a)~adding/removing JEPA in a recipe with \emph{no} oracle teacher (A6); (b)~adding/removing JEPA in a recipe with an \emph{unbiased} on-policy teacher.
Both are listed in \S\ref{sec:conclusion}; until they are completed, the ``buffer'' interpretation remains a hypothesis.

We therefore revise the positioning of util/rank: under the current regime, they should be treated as an optional, candidate-for-removal auxiliary rather than a default additive component.
Whether they remain net-negative under other oracle definitions, larger scales, or joint backbone training is an open question.

\subsection{Retrospective Diagnosis of the Oracle Design}
\label{sec:discuss:retro}

The ``subsequent-all-full'' oracle introduced in~\S\ref{sec:methods:util} was \emph{not} a speculative design choice selected in hindsight.
It was adopted because it is tractable, isolates the current-layer marginal value, and avoids coupling the teacher to an evolving gate policy---criteria that make it a reasonable first-version label when an explicit teacher is desired.
The A3 result (\S\ref{sec:results:a3}) is the post-hoc observation that the same property making the oracle tractable---assuming \emph{all} subsequent layers execute \texttt{full}---also makes it structurally off-policy with respect to actual gated execution, which at each subsequent controlled layer routes only a fraction $\rho$ of tokens through \texttt{full}.
Under this asymmetry the teacher can \emph{systematically underestimate} the relative cost of executing \texttt{cheap}, and under controller-only training the distributional distance between teacher and student tends to grow over 20k steps rather than fluctuate randomly (\S\ref{sec:methods:util}).
We therefore frame util/rank as a \textbf{reasonable initial design whose net effect turns out to be negative under the current recipe, with a structurally grounded but not yet causally demonstrated explanation}.
Concretely: (H1)/(H5)---off-policy / oracle-definition-driven mismatch---are consistent with this diagnosis; (H3)---a too-large weight---remains a simpler alternative that our data cannot exclude; the two explanations are not mutually exclusive.
The conceptual analogy to training--inference mismatch in sequential prediction~\citep{bengio2015scheduled} and to covariate shift in imitation learning~\citep{ross2011dagger} supports treating the mismatch as a plausible driver, but we do \emph{not} claim that these works prove the A3 mechanism.

\subsection{Connection to Related Work}
\label{sec:discuss:related}

The observation that auxiliary losses can produce a net-negative effect on LM quality in routing training finds complementary support in the MoE literature.
Router z-loss~\citep{zoph2022stmoe} and load-balancing loss~\citep{fedus2022switch} are \emph{numerical stability} constraints rather than \emph{policy quality} constraints.
Our utility/rank losses belong to the latter category: they attempt to directly specify ``which token is worth computing.''
Our experiments suggest that when the LM loss already provides on-policy gradients and the hard top-$k$ routing is insensitive to absolute score scale, such direct specification---under the current weights and oracle definition---may actually restrict the optimization space.
This distinction suggests that future router auxiliary loss design should \textbf{differentiate ``stability-oriented'' from ``policy-quality-oriented'' auxiliaries}: the former are often harmless and necessary; the latter require more careful verification that the teacher definition is consistent with on-policy execution and that the weight is appropriately calibrated.
A complementary line of work in the MoE literature pushes even further, advocating \emph{auxiliary-loss-free} load balancing so that routing gradients come exclusively from the task loss~\citep{wang2024auxfree}; while their setting is expert routing rather than depth routing, the underlying concern---that explicit routing auxiliaries may introduce undesired gradients---is structurally related to our A3 observation.

\subsection{Gradient Norms Require Joint Interpretation with Diagnostics}
\label{sec:discuss:grad}

\begin{figure}[t]
  \centering
  \includegraphics[width=\textwidth]{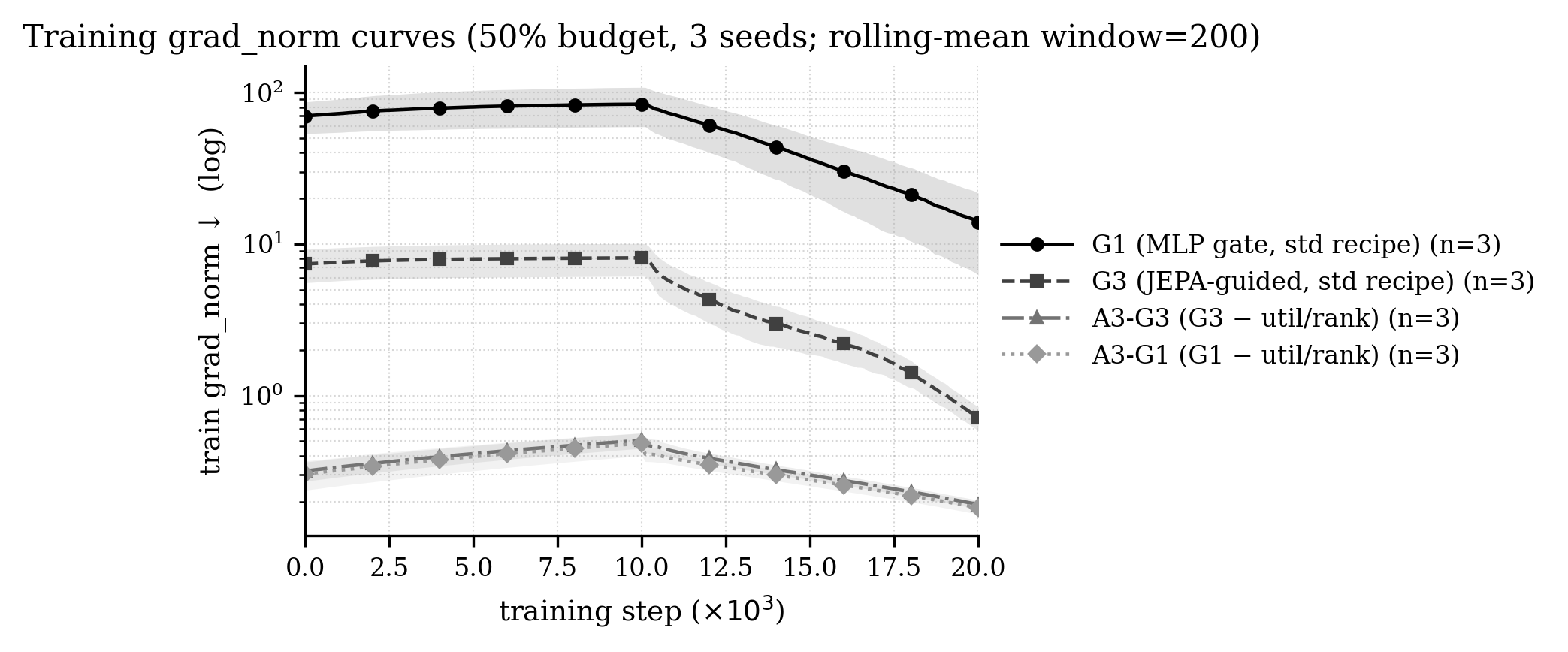}
  \caption{Training \texttt{grad\_norm} curves for four core configurations (log-$y$; solid = mean, shaded = $\pm 1$ std; rolling-mean window = 200 steps).
  Three tiers are clearly separated: G1 ${\sim}\,80$, G3 ${\sim}\,8$, A3-G3/A3-G1 ${\sim}\,0.5$.
  A1 is excluded from this figure because its low gradient norm (${\sim}\,2.3$) arises from predictor collapse (\S\ref{sec:results:a1}), qualitatively different from the ``lower norms with better LM and non-collapsed diagnostics'' shown here.}
  \label{fig:grad_norm}
\end{figure}

As noted in \S\ref{sec:results:a1}, A1 has the lowest \texttt{grad\_norm}---a symptom of predictor collapse, not healthier optimization.
Similarly, A3's gradient norm drops from G3's 8.06 to 0.51, concurrent with LM improvement and non-collapsed diagnostics---this \emph{is} genuinely ``cleaner optimization.''
\textbf{Relying on gradient norms as a standalone metric is misleading}; they become meaningful only when interpreted jointly with LM metrics, predictor diagnostics, and gate behavior.

\subsection{Limitations}
\label{sec:discuss:limits}

\begin{itemize}[leftmargin=*,itemsep=2pt]
\item \textbf{Scale.}
All experiments are at ${\approx}\,157.5$M total / ${\approx}\,0.83$--$0.85$M trainable parameters, controller-only.
Whether the gradient chain from LM loss to gate---which grows with depth---renews the value of explicit score teachers at larger scale is an open question.

\item \textbf{Budget range.}
Only 50\% (main) and 25\% (supplement) budget points.
The G3 vs.\ G1 endpoint LM direction is non-monotonic across $\rho$ (both within 0.005, but paired signs reverse at 25\%), suggesting budget dependence may be more complex than initially assumed.

\item \textbf{$\lambda_{\text{util}}$ / $\lambda_{\text{rank}}$ not swept.}
A4 sweeps only $\lambda_{\text{JEPA}}$.
The A3 finding therefore cannot be distinguished from the explanation that $(1.0, 0.2)$ happens to lie in an ``excessively strong'' region (H3).
This is the most direct methodological gap.

\item \textbf{\texttt{G1-costmatch} single seed.}
Only seed 42 completed; seeds 123 and 7 remain pending.

\item \textbf{No external benchmarks.}
Evaluation is limited to \texttt{eval lm\_loss} and derived metrics; downstream task validation is left to future work.

\item \textbf{A2 cross-ablation not conducted.}
A2 retains util/rank, so the separate contribution of semantic pairing vs.\ util/rank compensation cannot be cleanly isolated.

\item \textbf{Compute cost is proxy only.}
All cost figures are FFN-equivalent token-layer proxies; no real latency/throughput measurements are included.

\item \textbf{Oracle definition not varied.}
Only ``subsequent all full'' was tested; whether on-policy or TD-style utility would also yield net-negative results is unknown.
\end{itemize}

\section{Conclusion and Future Work}
\label{sec:conclusion}

This paper evaluates a JEPA-inspired gate for conditional depth routing and, through systematic ablation, discovers a recipe-level interaction between two auxiliary-loss families.
All experiments run on a single controlled platform (${\approx}\,157.5$M total parameters, ${\approx}\,0.83$--$0.85$M trainable controller parameters with a supplementary \texttt{G1-costmatch} at ${\approx}\,0.90$M; controller-only, 50\% budget).

\paragraph{Main conclusions} (from the 50\% budget, 3-seed main experiment unless otherwise noted):

\begin{enumerate}[leftmargin=*,itemsep=3pt]
\item \textbf{G3 improves optimisation dynamics under the standard recipe.}
The JEPA-guided gate achieves lower avg LM, faster threshold hits, and ${\approx}\,10.3{\times}$ lower gradient norms relative to the MLP gate in 3/3 seeds.
The 20k-step endpoint LM falls within the 0.005 heuristic reference threshold at 50\% budget; the 25\% budget supplement (2 seeds, lower evidence level) shows the same direction with per-seed sign reversal.
JEPA alignment is necessary for predictor non-collapse (A1) and weight-insensitive across $\{0.25, 1.0, 2.0\}$ (A4); $\ell_2$ distance rather than cosine similarity is the appropriate collapse diagnostic for action-conditional predictors in this setup.

\item \textbf{Removing util/rank improves LM and reveals G1--G3 equivalence (A3).}
Under the current recipe ($\lambda_{\text{util}} = 1.0$, $\lambda_{\text{rank}} = 0.2$, ``subsequent-all-full'' oracle), simultaneously removing utility regression and rank supervision:
(a)~consistently improves best/avg/hit metrics in 3/3 seeds for both gates;
(b)~collapses the A3-G1 vs.\ A3-G3 per-seed best eval lm difference to $|\Delta| \in [0.0004, 0.0012]$, all $< 0.005$;
(c)~reduces the training-time FLOPs proxy from ${\approx}\,1.53{\times}$ to ${\approx}\,1.07{\times}$ (measured wall-clock: ${\approx}\,39\%$ reduction).
The off-policy teacher bias (\S\ref{sec:methods:util}) is the structurally grounded working hypothesis; weight choice (H3) remains an alternative our data cannot exclude.
\end{enumerate}

\paragraph{Coverage of the JEPA $\times$ util/rank matrix.}
Viewing JEPA architecture (on/off) and util/rank (on/off) as two dimensions, this paper covers 3.5 of the 4 cells:

\begin{table}[h]
\centering
\small
\begin{tabular}{@{}lcc@{}}
\toprule
 & \textbf{util/rank: on} & \textbf{util/rank: off} \\
\midrule
\textbf{JEPA: on} & G3 (\S\ref{sec:results:main}) & A3-G3 (\S\ref{sec:results:a3}) \\
\textbf{JEPA: off} & A1 / G1 (\S\ref{sec:results:a1}, \ref{sec:results:main}) & A3-G1 (\S\ref{sec:results:a3}) \\
\bottomrule
\end{tabular}
\end{table}

JEPA's independent LM contribution (A3-G3 vs.\ A3-G1) is not detectable ($|\Delta| < 0.005$); its independent \emph{representation} contribution is confirmed (A1 collapse vs.\ non-collapse).

\paragraph{Future work} (ordered by directness to the main findings):

\begin{itemize}[leftmargin=*,itemsep=2pt]
\item \textbf{$\lambda_{\text{util}}$ / $\lambda_{\text{rank}}$ sweep.}
Sweep $\lambda_{\text{util}} \in \{0, 0.1, 0.25, 0.5, 1.0\}$ at fixed $\lambda_{\text{rank}} = 0.2$, and vice versa.
This is the most critical missing experiment: it determines whether A3 reflects inherent harm (H1/H2) or merely excessive weight (H3).

\item \textbf{On-policy utility.}
Replace the off-policy oracle with utility computed under the actual gate policy, testing whether the distributional mismatch is the primary driver of A3.

\item \textbf{Cross-ablation A5.}
Shuffled-target $+$ no-util/rank, isolating semantic pairing from util/rank compensation.

\item \textbf{Isolated predictor value A6.}
A1 $+$ no-util/rank, testing whether the predictor architecture provides any LM benefit without any oracle teacher (verifying the ``buffer hypothesis'' from \S\ref{sec:discuss:roles}).

\item \textbf{Util vs.\ rank decomposition.}
Retain only util or only rank to determine whether the net-negative contribution is driven by Huber regression, ranking, or their interaction.

\item \textbf{Budget sweep.}
$\rho \in \{0.10, 0.25, 0.50, 0.75\}$, confirming whether the non-monotonic G3 vs.\ G1 endpoint LM direction is systematic.

\item \textbf{Scale extension.}
Replicate key comparisons at 0.5B / 1B parameters.

\item \textbf{Joint training.}
Unfreeze the backbone; test whether gate and representation co-optimization alters the findings.
\end{itemize}

\paragraph{Stance.}
This paper contributes reproducible empirical observations about (a)~a JEPA-inspired gate design for conditional depth routing and (b)~auxiliary-loss interaction in that setting.
We do not advocate any particular gate training configuration as a recommended default; the finding that util/rank is net-negative should be treated as a \textbf{recipe-level working hypothesis awaiting further verification} via the experiments listed above.

\bibliographystyle{plainnat}
\bibliography{references}

\end{document}